\documentclass[a4paper,twoside]{article}

\usepackage{graphicx}
\usepackage{epsfig}
\usepackage{calc}
\usepackage{amssymb}
\usepackage{amstext}
\usepackage{amsmath}
\usepackage{amsthm}
\usepackage{multicol}
\usepackage{pslatex}
\usepackage{apalike}
\usepackage{algpseudocode}
\usepackage{amsfonts}
\usepackage{amsmath}
\usepackage{caption}
\usepackage{subfigure}
\usepackage{SCITEPRESS}
\newcommand{\quotes}[1]{``#1''}

\begin{document}

\title{AMSOM: Adaptive Moving Self-organizing Map \\for Clustering and Visualization}

\author{\authorname{Gerasimos Spanakis\sup{1} and Gerhard Weiss\sup{1}}
\affiliation{\sup{1}Department of Knowledge Engineering, Maastricht University, 6200MD, Maastricht, The Netherlands}
\email{\{jerry.spanakis, gerhard.weiss\}@maastrichtuniversity.nl}}


\abstract{Self-Organizing Map (SOM) is a neural network model which is used to obtain a topology-preserving mapping from the (usually high dimensional) input/feature space to an output/map space of fewer dimensions (usually two or three in order to facilitate visualization). Neurons in the output space are connected with each other but this structure remains fixed throughout training and learning is achieved through the updating of neuron reference vectors in feature space. Despite the fact that growing variants of SOM overcome the fixed structure limitation they increase computational cost and also do not allow the removal of a neuron after its introduction. In this paper, a variant of SOM is proposed called AMSOM (Adaptive Moving Self-Organizing Map) that on the one hand creates a more flexible structure where neuron positions are dynamically altered during training and on the other hand tackles the drawback of having a predefined grid by allowing neuron addition and/or removal during training. Experiments using multiple literature datasets show that the proposed method improves training performance of SOM, leads to a better visualization of the input dataset and provides a framework for determining the optimal number and structure of neurons.}

\onecolumn \maketitle \normalsize \vfill

\section{\uppercase{Introduction}}
The Self-Organizing Map (SOM) \cite{kohonen2001self} is an unsupervised neural network model which effectively maps high-dimensional data to a low-dimensional space (usually two-dimensional). The low-dimensional space (also called output space) consists of a grid of neurons connected with each other, according to a specific structure (can be hexagonal, rectangular, etc.). This structure allows the topology preservation of input data (i.e., similar input patterns are expected to be mapped to neighboring neurons in the output grid) \cite{kohonen1981automatic}. By this way, SOM manages to achieve dimensionality reduction, abstraction, clustering and visualization of the input data and this is the reason that it has been applied successfully to many different domains and datasets like financial data \cite{deboeck2013visual}, speech recognition \cite{kohonen1988neural}, image classification \cite{lu1990pattern}, document clustering \cite{lagus1999websom}, \cite{spanakis2012doso}  .

The SOM algorithm raises some issues and problems: (1) SOM's architecture is fixed and predefined in terms of number and arrangement of neurons. In case of largely unknown input data, it is difficult to determine apriori the correct structure that provides satisfactory results. There is some work in this area in order to how to add/remove neurons but none of current approaches adjusts neuron positions on the grid according to training progress. (2) Training a SOM comes with a large computation cost, especially in cases of large datasets and/or large maps. Many epochs might be needed in order for the SOM to converge and the map to reach a final state. 

In this paper we propose an extension of the traditional SOM, which handles both issues described above: First, it allows neurons to change positions during training which provides better visualization and faster training time. Second, number of neurons can be adjusted (neurons can be either added or removed) according to dataset requirements and training progress. Due to this enhanced training scheme, the number of epochs required for training is significantly reduced. The rest of the paper is organized as follows. Section \ref{sect:related} presents background work on SOM, extensions on the traditional algorithm and their limitations. The proposed method is presented in Section \ref{sect:soc} while experimental setup is described in Section \ref{sect:exp}. Finally, Section \ref{sect:concl} concludes the paper.

\section{\uppercase{Related work}}
\label{sect:related}
\subsection{SOM and Competitive Learning}
The Self-Organizing Map (SOM) is a fully connected single-layer linear neural network. The SOM uses a set of neurons, often arranged in a 2-D rectangular or hexagonal grid, to form a discrete topological mapping of an input space, $\mathbf{X} \in R^D$. Input space consists of a set of vectors $\mathbf{x}_{j} \in R^D$:

\begin{equation}
\mathbf{x}_j=[x_{j1},x_{j2},...,x_{jD}]^T
\end{equation}

$\mathbf{w}_{i}$ is the weight vector associated to neuron $i$ and is a vector of the same dimension ($D$) of the input space, $M$ is the total number of neurons. Obviously, these weights represent the synaptic connections of each neuron $i$ and can be denoted:

\begin{equation}
\mathbf{w}_{i}=[w_{i1},w_{i2},...,w_{iD}]^T
\label{eqn:somw}
\end{equation}

The fundamental principle of SOM is the soft competition between the nodes in the output layer; not only the node (winner) but also its neighbors are updated \cite{kohonen2012self}.

\begin{figure}[htb]
\centering
  \includegraphics[width=1.0\linewidth]{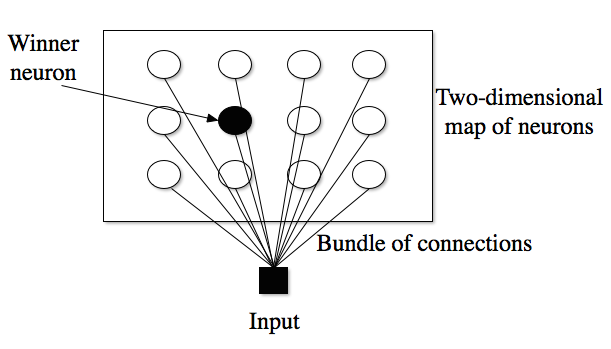}
  \caption{The SOM (fully connected) architecture.}
  \label{fig:somarch}
\end{figure}

A SOM architecture can be found in Figure \ref{fig:somarch}.

All the weights ${\mathbf{w}_{1},\mathbf{w}_{2},...,\mathbf{w}_{M}}$ are initialized to random numbers, in the range of the corresponding input characteristics. We also introduce a discrete time index $t$ such that $\mathbf{x}(t), t=0,1,...$ is presented to network at time $t$ and $\mathbf{w}_{i}(t)$ is the weight vector of neuron $i$ computed at time $t$. The available input vectors are recycled during the training (or learning) process: a single pass over the input data is called an epoch.

\subsubsection{On-line Training of SOM}
In the conventional \quotes{on-line} or \quotes{flow-through} method, the weight vectors are updated recursively after the presentation of each input vector. As each input vector is presented, the Euclidean distance between the input vector and each weight vector is computed:

\begin{equation}
d_i(t)=||\mathbf{x}(t)-\mathbf{w}_i(t)||^2
\end{equation}

Next, the winning or best-matching node (denoted by subscript $c$) is determined by:

\begin{equation}
c = \{i, min_{i}d_{i}(t)\}
\end{equation}

Note that we suppress the implicit dependence of $c$ on discrete time $t$. The weight vectors are updated using the following rule:

\begin{equation}
\mathbf{w}_{i}(t+1)=\mathbf{w}_{i}(t) + \alpha(t) \cdot h_{ci}(t) \cdot \left [ \mathbf{x}(t)-\mathbf{w}_{i}(t)  \right ]
\label{eqn:somupdate}
\end{equation}

where $\alpha(t)$ is the learning-rate factor and $h_{ci}(t)$ is the neighborhood function. The learning rate factor controls the overall magnitude of the correction to the weight vectors, and is reduced monotonically during the training phase. The neighborhood function controls the extent to which $\mathbf{w}_i(t)$ is allowed to adjust in response to an input most closely resembling $\mathbf{w}_c(t)$ and is typically a decreasing function of the distance on the 2-D lattice between nodes $c$ and $i$. We use the standard Gaussian neighborhood function:

\begin{equation}
h_{ci}(t)=exp \left( -\frac{||\mathbf{r}_i-\mathbf{r}_c||^2}{\sigma(t)^2} \right )
\label{eqn:somhc}
\end{equation}

where $\mathbf{r}_i$ and $\mathbf{r}_c$ denote the coordinates of nodes $i$ and $c$, respectively, on the output space (usually two-dimensional grid). The width $\sigma(t)$ of the neighborhood function decreases during training, from an initial value comparable to the dimension of the lattice to a final value effectively equal to the width of a single cell. It is this procedure which produces the self-organization and topology preserving capabilities of the SOM: presentation of each input vector adjusts the weight vector of the winning node along with those of its topological neighbors to more closely resemble the input vector. The converged weight vectors approximate the input probability distribution function, and can be viewed as prototypes representing the input data.

\subsubsection{Batch Training of SOM}
The SOM update given by Equation (\ref{eqn:somupdate}) is \quotes{on-line} in the sense that the weight vectors are updated after the presentation of each input record. In the batch SOM algorithm (proposed in \cite{kohonen1993things}), the weights are updated only at the end of each epoch according to:

\begin{equation}
\mathbf{w}_{i}(t_f)=\frac{\sum_{t'=t_0}^{t'=t_{f}}\tilde{h}_{ci}(t') \cdot \mathbf{x}(t')}{\sum_{t'=t_0}^{t'=t_{f}}\tilde{h}_{ci}(t')}
\label{som:batch1}
\end{equation}

where $t_0$ and $t_f$ denote the start and finish of the present epoch, respectively, and $w_{i}(t_f)$ are the weight vectors computed at the end of the present epoch. Hence, the summations are accumulated during one complete pass over the input data. The winning node at each presentation of new input vector is computed using:

\begin{eqnarray}
\tilde{d}_{i}(t)=||\mathbf{x}(t)-\mathbf{w}_{i}(t_0)||^2\\
c = \{i, min_{i}\tilde{d}_{i}(t)\}
\label{eqn:batchwinner}
\end{eqnarray}

where $\mathbf{w}_i(t_0)$ are the weight vectors computed at the end of the previous epoch. The neighborhood functions $\tilde{h}_{ci}(t)$ are computed using Equation (\ref{eqn:somhc}), but with the winning nodes determined from Equation (\ref{eqn:batchwinner}). This procedure for computing the neighborhood function is identical to the Voronoi partinioning. As is in the on-line method, the width of the neighborhood function decreases monotonically over the training phase.

A more concrete explanation of the batch algorithm is given by the following Equation:

\begin{equation}
\mathbf{w}_{i}=\frac{\sum_{j} n_j \cdot h_{ji} \cdot \tilde{x}_{j}}{\sum_{j} n_j \cdot h_{ji}}
\label{som:batch2}
\end{equation}

where $n_j$ is the number of input items mapped into node $j$ and the index $j$ runs over the nodes in the neighborhood of node $i$. The basic idea is that for every node $j$ in the grid, the average $\tilde{x}_{j}$ of all those input items $x(t)$ is formed that have node $j$ (i.e., vector $\mathbf{w}_{j}$) as the closest node. The above Equation is used for updating the node weight vectors and this is repeated for a few times, always using the same batch of input data items to determine the updated $\tilde{x}_{j}$.

The batch SOM offers several advantages over the conventional on-line SOM method. Since the weight updates are not recursive, there is no dependence upon the order in which the input records are presented. In addition to facilitating the development of data-partitioned parallel methods, this also eliminates concerns \cite{mulier1994learning} that input records encountered later in the training sequence may overly influence the final results. The learning rate parameter $\alpha(t)$ does not appear in the batch SOM algorithm, thus eliminating a potential source of poor convergence \cite{ceccarelli1993competitive} if this parameter is not properly specified.

The mathematical theory of the SOM is very complicated and only the one-dimensional case has been analyzed completely \cite{fort2006som}, since the SOM belongs to the \lq ill posed\rq\ problems in mathematics. The SOM can also be looked at as a \lq nonlinear projection\rq\ of the probability density function of high-dimensional input data onto the two-dimensional display. 

\begin{figure}[htb]
  \centering
  \subfigure[Hexagonal grid]{
        \includegraphics[width=0.48\columnwidth,keepaspectratio]{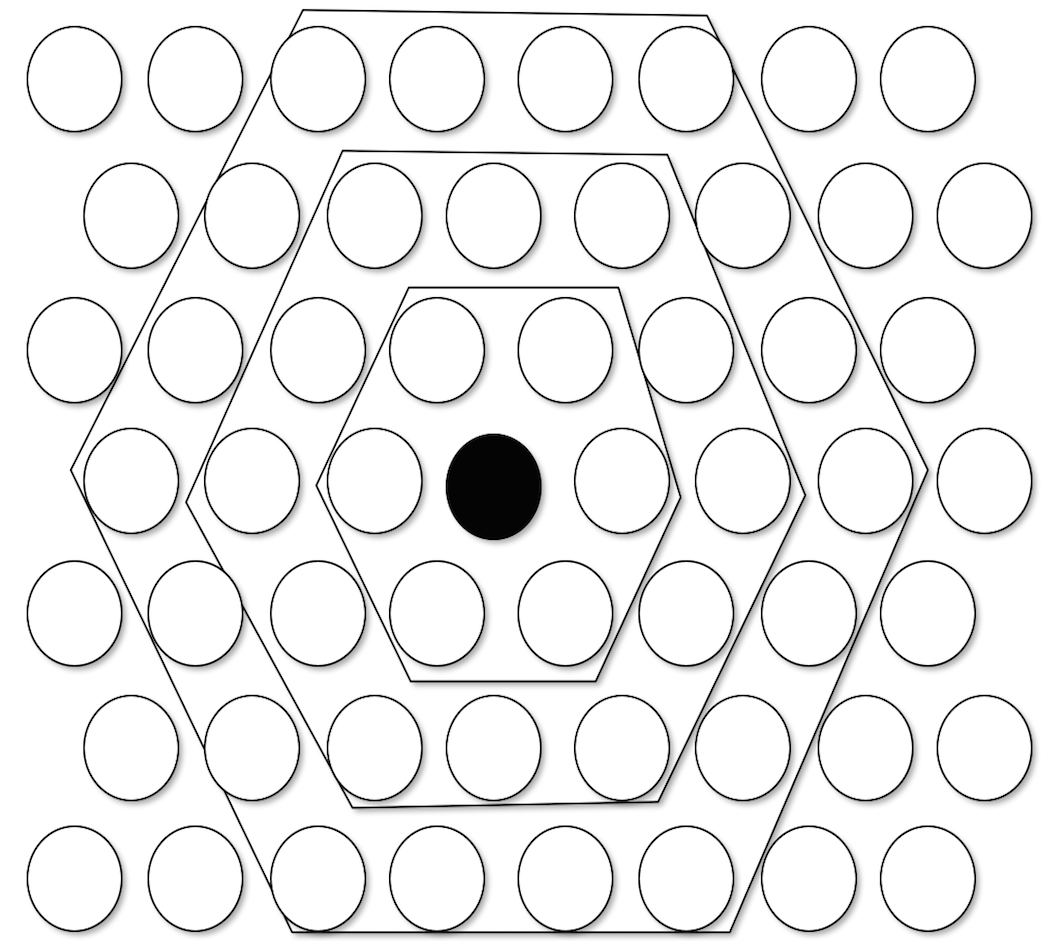}}
  \subfigure[Rectangular grid]{
        \includegraphics[width=0.48\columnwidth,keepaspectratio]{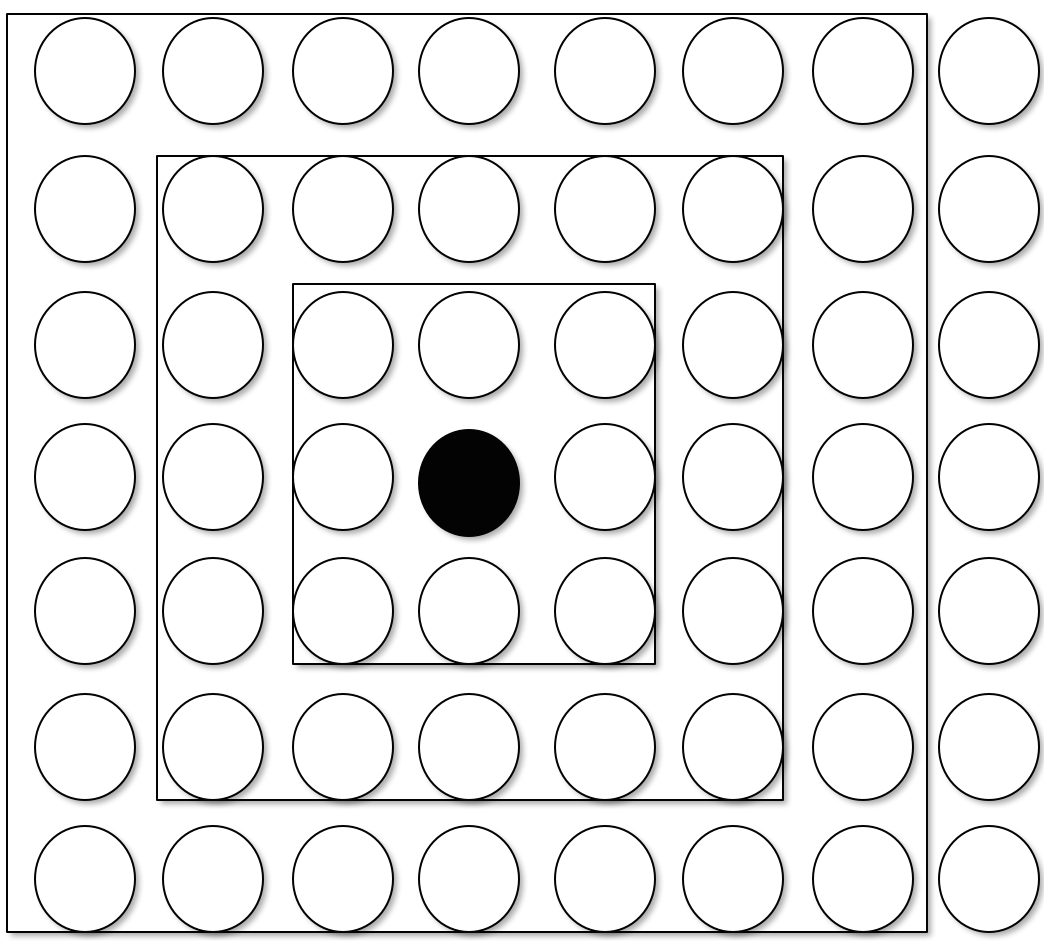}}
\caption{Typical SOM grids with different neighborhoods around the winner neuron.}
\label{fig:somgrid}
\end{figure}

Usually, the input is mapped onto a 1- or 2-dimensional map. Mapping onto higher dimensions is possible as well, but complicates the visualization. The neurons connected to adjacent neurons by a neighborhood relationship define the structure of the map. The two most common 2-dimensional grids are the hexagonal grid and the rectangular grid and are shown in Figure \ref{fig:somgrid}.

The neighborhood function defines the correlation between neurons. The simplest neighborhood function is called \textit{bubble}; it is constant over the neighborhood of the winner neuron and zero otherwise. The neighborhood of different sizes in rectangular and hexagonal maps can be seen in Figure \ref{fig:somgrid}. A more flexible definition is the \textit{gaussian neighborhood function} defined by Equation (\ref{eqn:somhc}).

The number of neurons, the dimensions of the map grid, the map lattice and shape must be specified before training. The more neurons the grid has, the more flexible the mapping becomes but the computation complexity of the training phase increases as well. The choice of the map structure and size is both related to the type of problem and the subjective choice of the user.

\subsection{Flexible Structure in Neural Networks and SOM}
The norm in artificial neural nets is that classic techniques involve simple and often fixed network topologies trained via stimulus-based methods such as backpropagation. However, there are cases in which the structural design of the network is strongly influenced by the environment and by utilizing constructive and pruning algorithms. Both these algorithmic categories deliver a network which is gradually adjusted in response to training data. There are many approaches which apply these algorithms in classic neural networks \cite{islam2009new}, \cite{bortman2009growing}, \cite{han2013structure}, \cite{yang2012evolutionary}. 

Also, there are many variations of SOM that allow a more flexible structure of the output map which can be divided into two categories: In the first type, we include growing grid (GG) \cite{fritzke1995gg}, incremental GG \cite{blackmore1993incremental}, growing SOM (GSOM) \cite{alahakoon2000dynamic} all coming with different variants. GG is the only variant which allows growing a new node from the interior of the grid (but this is a whole row or column of nodes). In the rest cases, new nodes are generated by a boundary node, despite the fact that the highest error could have been generated by an internal node. The idea is that the error will be propagated to the exterior to guarantee that growing can only be from the boundaries but this process can lead to a map structure with not perfect topology preservation. Therefore, map size becomes very wide after a limited number of insertions, with some additional nodes, which have no effect. MIGSOM \cite{ayadi2012migsom} allows a more flexible structure by adding neurons internally and from the boundary but still does not offer the ability to remove neurons if necessary.

In the second type of growing variants, the rectangular grid is replaced with some connected
nodes. We denote growing cell structures (GCSs) \cite{fritzke1994growing}, growing neural gas (GNG) \cite{fritzke1995growing} and growing where required \cite{marsland2002self}. These works add just the necessary nodes at the same time, to fine-tune the optimal map size. Nevertheless, GCS and GNG are facing many difficulties for visualizing high-dimensional data. Visualization in these cases is guaranteed only with low-dimensional data. 

Limitations in growing and visualization led to hierarchical variants of the previous model like the Growing Hierarchical SOM (GHSOM) \cite{rauber2002growing}. With GHSOM you can get an idea of the hierarchical structure of the map, but the growing parameter of the map has to be defined beforehand. Other approaches (like TreeGNG \cite{doherty2005treegng} or TreeGCS \cite{hodge2001hierarchical}) use dendrograms for representation but due to this tree structure they lose the topological properties.

Disadvantages of these approaches are: (a) the high computational cost due to the fact that structure starts from a very basic architecture and has to grow in order to reach an acceptable structure for the data and (b) the fact that after adding neurons there is not the possibility of removing a neuron if performance is not improving. 

\section{\uppercase{Expanding the idea of self-organization in neuron locations}}
\label{sect:soc}

During the classic SOM algorithm neuron positions remain unchanged and the grid is fixed from the beginning till the end of the training. This facilitates the process of learning (since neighborhood structure is known beforehand) but is restricting regarding the final result and ways of visualizing it. We propose a different and more flexible scheme in regard to position vectors $\mathbf{r}_{i}$ of neurons, which allows a more adaptive form of the neuron grid and acts as an extension to the batch learning algorithm.

Starting from an already grown map size, AMSOM can adapt both its size and structure in order to better represent the data at a specific level of detail. After a specific number of steps, neurons are analyzed to see whether the level of representation is sufficient or adjustments are needed: removal and/or addition of neurons. Initially, connections between neurons are determined based on the grid structure but as training advances, these can change and adjust according to the way that neuron positions are also changed during the process. The algorithm flow is described in Figure \ref{fig:algo} and more details about the steps are presented in the following subsections.

\begin{figure*}[htb]
\centering
\begin{algorithmic}
\State \textbf{1. Initialization Phase}
\State 1.1: Derive initial grid structure and size (number of neurons $M$) of the AMSOM
\State 1.2: Initialize weight vectors ($\mathbf{w}_i$) to random values (according to the value range of features).
\State 1.3: Initialize position vectors ($\mathbf{r}_i$) according to the initial grid structure
\State 1.4: Initialize edge connectivity matrix ($\mathbf{E}$) values according to the grid connections
\State 1.5: Initialize edge age matrix ($\mathbf{A}$) values to zero
\State 1.6: Define growing threshold ($GT$) according to dimension of the data $D$ and a spreading factor ($SF$).
\State \textbf{2. Training phase}
\For {$t=1:maxepochs$}
\For {$i=1:P$}
\State 2.1: Find winner neuron $N_a$ according to Equation (\ref{eqn:batchwinner}) and increase times that neuron $N_a$ is winner by 1
\State 2.2: Find second best matching neuron $N_b$ (using Equation (\ref{eqn:batchwinner}) and excluding $N_a$ from the search)
\State 2.3: Age of all edges between $N_a$ and its neighbors increased by one
\State 2.4: Connect $N_a$ with $N_b$ (if they were not already connected)
\State 2.5: Reset age between $N_a$ and $N_b$ to zero
\EndFor
\State 2.6: Use Equations \ref{eqn:msomtrain}-\ref{eqn:msomtrainb} to update neuron weights.
\State 2.7: Use Equations \ref{eqn:posupd}-\ref{eqn:posupdb} to update neuron positions.
\State 2.8: \If {neurons need to be added/removed (check $age_{max}$ and $t_{add}$)} add/remove neurons and update accordingly
\EndIf
\State 2.9: \If {error does not change significantly} end training phase
\Else {Continue}
\EndIf
\EndFor
\State \textbf{3. Smoothing phase}
\State 3.1: Fine-tune weights and deliver the AMSOM neuron weight vectors and positions
\end{algorithmic}
\caption{AMSOM algorithm overview.}
\label{fig:algo}
\end{figure*}

\subsection{Phase I: AMSOM Initialization}
\subsubsection{Grid Structure and Size.}
The first step of AMSOM algorithm is to define the initial grid structure (as the classic SOM). This process facilitates training time in contrast to starting from a small-size structure and building on that as other approaches do \cite{vesanto2000som}. It is also in agreement with the neural development which suggests that nearly all neural cells used through human lifetime have been produced in the first months of life \cite{dowling2007great}. This overproduction of neuron cells is thought to have evolved as a competitive strategy for the establishment of efficient connectivity \cite{Changeux}.

\begin{figure}
\centering
  \includegraphics[width=1.0\linewidth]{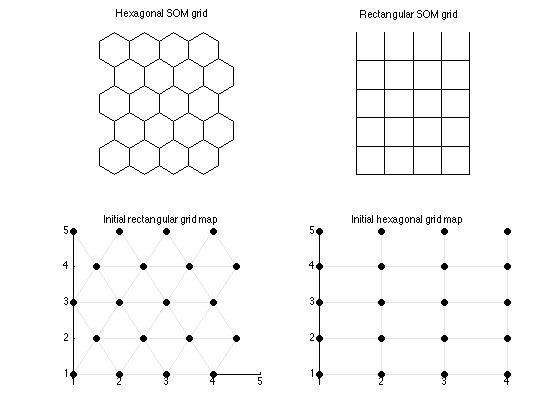}
  \caption{Initial grid example (hexagonal \& rectangular).}
  \label{fig:initgrid}
\end{figure}

Having this in mind, the initial structure of SOM is determined. Several empirical rules \cite{Park2006247} suggest that the number of neurons should be $5 \cdot \sqrt{N}$ where $N$ is the number of patterns in the dataset. In this case, the two largest eigenvalues of the training data are first calculated, then the ratio between side lengths of the map grid is set to the ratio between the two maximum eigenvalues. The actual side lengths are finally set so that their product is close to the number of map units determined according to \cite{vesanto2000som} rule. The eigenvalues ratio shows how well the data is flattened and elongated \cite{estevez2012advances}. At this point a more precise determination of the number of neurons is not essential, since this number will be fine tuned during the training process. Initially, neurons are connected with their neighbors following the idea of Figure \ref{fig:somgrid} using a rectangular or hexagonal grid. For example, if the algorithm suggests that the initial grid of the AMSOM should be 5x4 (let's suppose rectangular), every neuron has 4 neighbors (except the marginal ones). Figure \ref{fig:initgrid} demonstrates two different topologies, a rectangular and a hexagonal one with the corresponding connections between neurons.

\subsubsection{Vector, Matrix and Parameters Initialization.}
For each neuron the following are defined and initialized accordingly:
\begin{itemize}
\item Neuron vector (weight vector, $\mathbf{w}_i$): It is the same as the classic SOM (see Equation (\ref{eqn:somw})) and shows the representation of the neuron in the feature (input) space. Initialization of neuron vectors is random according to literature standards.
\item Neuron position (position vector, $\mathbf{r}_i$): Depending on the output space (mostly it is two-dimensional), it's a vector that shows the position of the neuron. Initial position vectors are equal to the positions of the neurons in the grid, i.e., in Figure \ref{fig:initgrid} one can see the coordinates of neurons according to the structure (hexagonal or rectangular).
\end{itemize}

\begin{figure}[htp]
\centering
\includegraphics[width=0.75\linewidth]{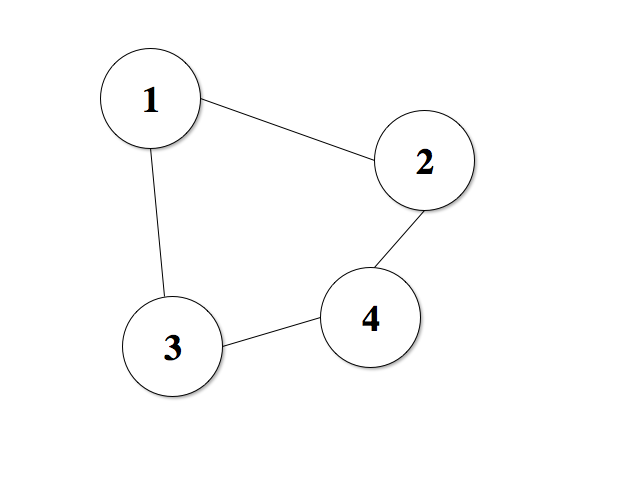}
\begin{equation*}
A=\begin{bmatrix}
0 & 2 & 4 & 0\\ 
2 & 0 & 0 & 0\\ 
4 & 0 & 0 & 1 \\
0 & 0 & 1 & 0
\end{bmatrix}
,
E=\begin{bmatrix}
0 & 1 & 1 & 0\\ 
1 & 0 & 0 & 1\\ 
1 & 0 & 0 & 1 \\
0 & 1 & 1 & 0
\end{bmatrix}
\end{equation*}
\caption{Example of matrices $A$ and $E$ describing connections between AMSOM neurons.}
\label{fig:matrices}
\end{figure}

Since the structure of the grid is subject to changes during training, we need to keep track of the neighbors of each neuron. There is the possibility that some neurons which where connected in the initial grid become disconnected after some time (or vice versa). In order to keep track of these changes we introduce the orthogonal and symmetrical matrices $\mathbf{E}$ and $\mathbf{A}$ (both size $M \times M$) where $E(p,q)$ shows if neurons $p$ and $q$ are connected (0 translates to no connection, 1 translates to connected neurons) and $A(p,q)$ shows the age of edge (as implied by $E(p,q)$) between neurons $p$ and $q$: This will be used in order to determine which neurons had incidental connections to other neurons or strong connections as training moves forward. When $A(p,q)$ is 0 that means that neurons $p$ and $q$ were closest neighbors at current epoch but any other value (i.e., 2) implies that neurons $p$ and $q$ were closest neighbors some epochs before (i.e., 2). An example of matrices $\mathbf{E}$ and $\mathbf{A}$ is seen in Figure \ref{fig:matrices}.

In this example, neurons number ($M$) is 4 and connectivity matrix $E$ shows how neurons are connected to each other (as implied by the graph). Age matrix $A$ shows how many epochs an edge has \lq survived\rq: Connection between neuron $\#1$ and $\#2$ has age 2 whereas connection between neuron $\#2$ and $\#4$ has age 0. Notice that age 0 can either mean that neurons are not connected, like neurons $\#1$ and $\#4$ or that neurons are connected at this current epoch (so their connection is \lq recent\rq), like neurons $\#2$ and $\#4$. 

Also, at this stage the growing threshold $GT$ of the map is defined as a function of data dimension ($D$) and a spread factor ($SF$) defined by the user. Formula used is $GT=-ln(D) \times ln(SF)$ (from \cite{alahakoon2000dynamic}). Generally, a $SF$ value of $0.5$ always yields good results but its fine tuning is up to the user requirements and the dataset structure.

\subsection{Phase II: Training}
\subsubsection{Weight and Position Updating}
For the weight learning of neurons, the SOM batch algorithm is utilized, as it was given in Equations \ref{som:batch1}-\ref{som:batch2}, which are repeated here for clarity. 

\begin{equation}
\mathbf{w}_{i}(t+1)=\frac{\sum_{j} n_j(t) \cdot h_{ji}(t) \cdot \mathbf{\tilde{x}}_{j}(t)}{\sum_{j} n_j(t) \cdot h_{ji}(t)}
\label{eqn:msomtrain}
\end{equation}
\begin{equation}
h_{ji}(t)=exp \left( -\frac{||\mathbf{r}_j-\mathbf{r}_i||^2}{\sigma(t)^2} \right )
\label{eqn:msomtrainb}
\end{equation}
where:
\begin{itemize}
\item $\mathbf{w}_i(t+1)$ marks neurons $i$ updated weight (at epoch $t+1$),
\item $t$ marks current epoch and $t+1$ marks the next epoch,
\item $n_j(t)$ marks the number of patterns that are assigned to neuron $j$,
\item $h_{ji}(t)$ marks the neighborhood function and is a measure of how close are neuron $j$ and neuron $i$,
\item $\mathbf{\tilde{x}}_j(t)$ is the mean feature vector of all $x$ that are assigned to neuron $j$ at epoch $t$,
\item $\mathbf{r}_j,\mathbf{r}_i$ are the position vectors (in the output space) for neurons $j$ and $i$,
\item $\sigma(t)$ is the adaptation factor, decreasing through training
\end{itemize}

Building on top of this, at the end of each epoch, the neuron position vectors are adjusted in a similar manner to the SOM training algorithm. In more detail, at the end of each epoch and after the neuron weight vectors update is over, the distances between the neuron vectors ($\mathbf{w}_{i}$) are computed. These distances show how close neurons are (in the input space) and can be used as a measure in order to update neuron positions (in the output space). This is achieved through the following Equations:

\begin{equation}
\mathbf{r}_{i}(t+1)=\mathbf{r}_{i}(t)+\alpha(t) \cdot \frac{\sum_{j}n_j(t) \cdot \delta_{ji}(t)(\mathbf{r}_{j}(t)-\mathbf{r}_{i}(t))}{\sum_{j} n_j(t) \cdot \delta_{ji}(t)}
\label{eqn:posupd}
\end{equation}
\begin{equation}
\delta_{ji}(t)=exp \left( -\frac{||\mathbf{w}_j-\mathbf{w}_i||^2}{\gamma \times \sigma(t)^2} \right )
\label{eqn:posupdb}
\end{equation}
where:
\begin{itemize}
\item $t$, $n_j(t)$ were defined in Equations \ref{eqn:msomtrain} and \ref{eqn:msomtrainb},
\item $\alpha(t)$ denotes the learning rate at epoch $t$ and controls the rate that positions of neurons are moving,
\item $\delta_{ji}(t)$ is a neighborhood function denoting how close neurons $j$ and $i$ are (during time $t$ and is based on their distance in the input space (i.e., distance computed based on their vectors $w_i$),
\item $\gamma$ is a parameter that controls the neighborhood shrinking as a fraction of $\sigma$ which was used in Equation (\ref{eqn:msomtrainb})
\end{itemize}

Notice the similarity of $\delta_{ji}$ with $h_{ji}$: both are neighborhood functions and are used to determine how close two neurons are but the first one does so using their distances in the feature (input) space while the latter does so using their distances in the output space (map). 

Equation (\ref{eqn:posupd}) will adjust neuron's $i$ position vector according to the neurons which proved winners for more patterns in its neighborhood and less (or even none) according to neurons which were winners for few patterns (or none). This process enhances the concept of neighborhood around the neurons that attract more patterns and also allows to cover any empty spaces in the data representation. It is expected to improve the training speed, since position updating will lead to more accurate position vectors that will be used for the next training epoch and leads to more insightful representations of the neurons in the output space.

Learning rate $\alpha(t)$ can also be set to a small value $0.01$ since the neighborhood function controls well the percentage of change in the position vectors. It was selected to update the position vectors with this hybrid on-line-batch SOM rule, due to the fact that output space is much smaller (in most SOM applications) than the input space, so in many cases minor adjustments (than major repositioning of the neurons) are more necessary in order to guarantee satisfactory training but also representation. Also note that the parameter $\gamma$ which controls neighborhood shrinking for position can also control how fast the map will be updated and how neurons are going to affect each other.

\subsubsection{Adding and Removing Neurons}
\begin{figure}[htb]
\centering
  \includegraphics[width=1.0\linewidth]{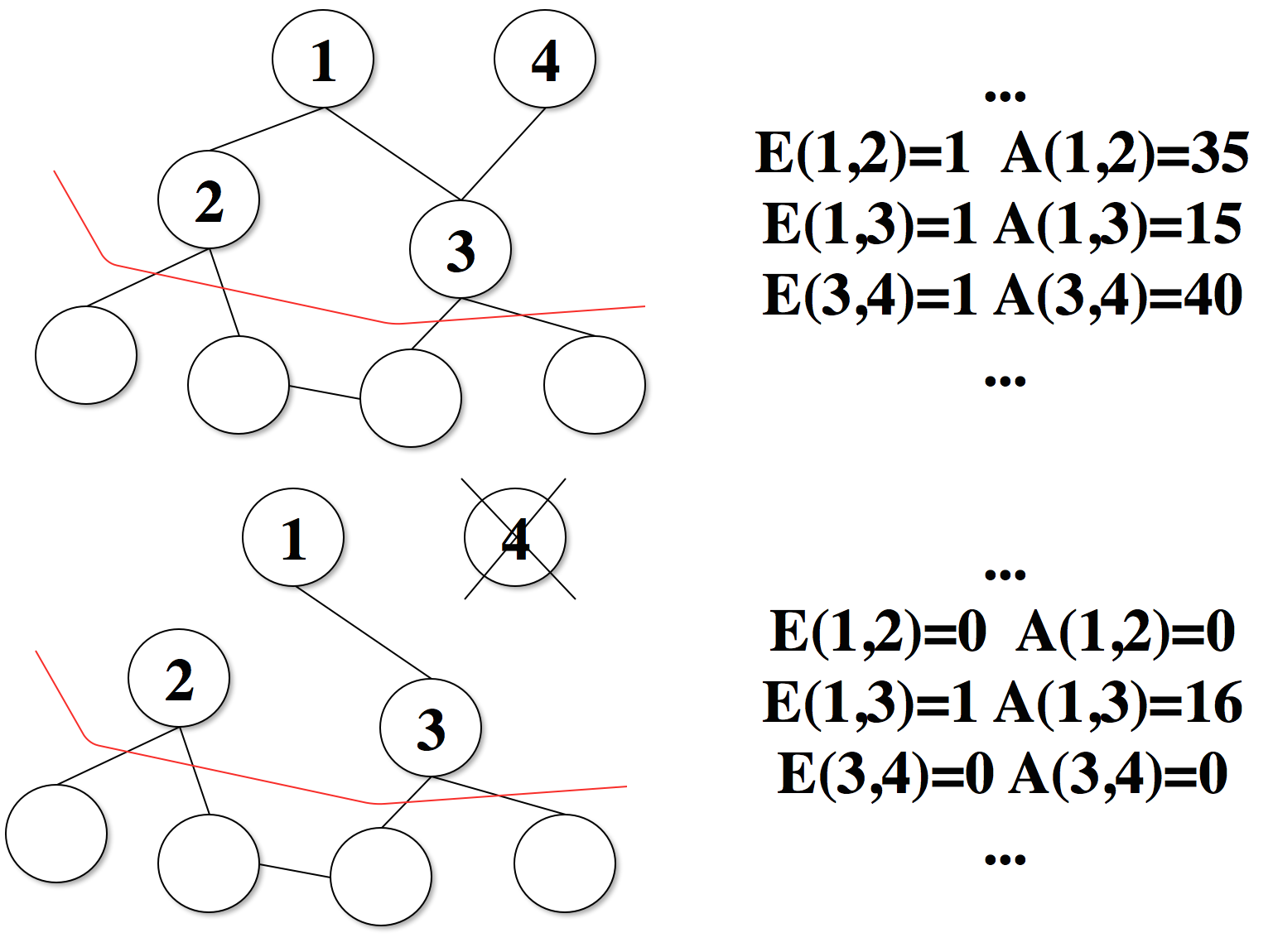}
  \caption{The process of removing neurons in a part of AMSOM: With $age_{max}$ set to 30, neuron 4 is disconnected from neuron 3 and neuron 1 is disconnected from neuron 2 (notice that matrices $A$ and $E$ are updated accordingly). Neuron 4 is left with no connections so it is removed.}
  \label{fig:msomrem}
\end{figure}

During the weight updating process, for each input (pattern) the best matching neuron is determined ($N_a$) and also the second best matching ($N_b$). At this step the age of all edges between $N_a$ and its neighbors is increased. Afterwards, $N_a$ is connected to $N_b$. If both of the neurons were already connected then their age is reset to zero. This is another step that implements the competitive learning rule, since for each new pattern, a new edge connecting the two closest neurons is drawn. This process is repeated for all patterns as they are presented to the AMSOM. Finally, at the end of the epoch for each incident edge between neurons ($i,j$), if $A(i,j) \geq age_{max}$, then this edge is removed. $age_{max}$ can be set to a value not small enough (so as to avoid many disconnections) but also not big enough (so as to avoid having a fully connected grid). In our experiments this value was 30. If either of the implicated neurons becomes isolated from the remainder of the structure, then it is removed from the grid. The aim here is to remove edges that are no longer useful because they are replaced by younger edges that are created during the AMSOM training. That is the reason that each time two neurons are connected by an edge, then its age is reset to zero. By this process, neurons that were connected incidentally -especially at the beginning of the training when the map is still under forming- are disconnected after some epochs. This process has two distinct advantages: (a) self-organization and competitive learning will allow after some epochs the removal of redundant number of neurons and (b) adjustment of connections between neurons so as to enhance topological properties of the dataset. An example of a removal of a neuron is shown in Figure \ref{fig:msomrem} along with the necessary adjustments to matrices $A$ and $E$.

Also, there is the possibility that after some epochs ($t_{add}$), new neurons are added. The criterion is based on the training progress and when an addition happens, then new neurons can be added only after a number of epochs ($t_{add}$) in order to allow weight adaptation of the map, before evaluating current structure. First step is to spot the neuron $N_u$ with the largest quantization error. A new neuron will be added, if its quantization error is higher than $GT$, where $GT$ is the growing threshold of the map: A high value for $GT$ will result in less spread out map and a low $GT$ will produce a more spread map. If the quantization error satisfies the above condition then its Voronoi region is considered to be under-represented and therefore a new neuron has to be added to share the load of the high-error-valued neuron.

\begin{figure}[htb]
\centering
  \includegraphics[width=1.0\linewidth]{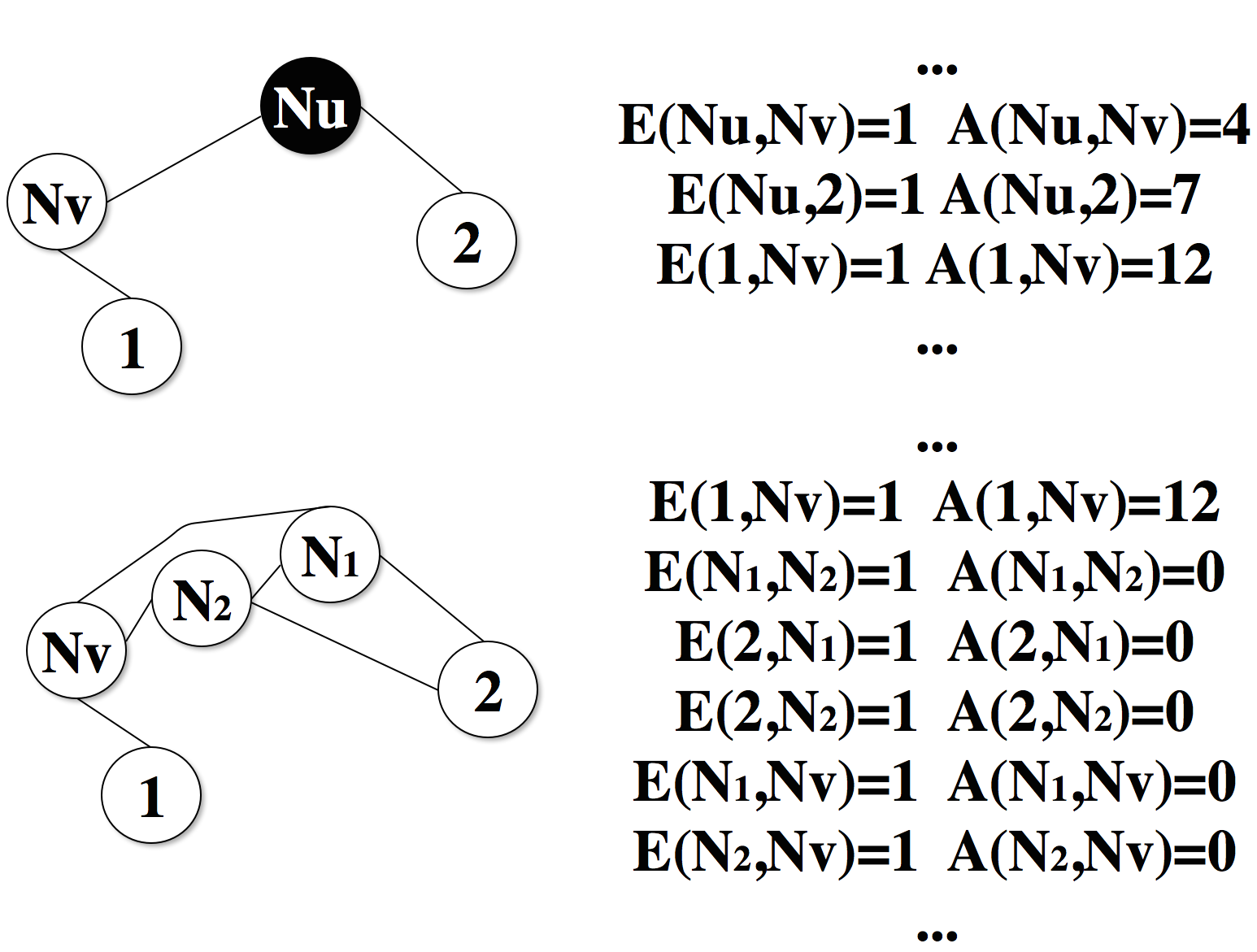}
  \caption{The process of adding new neurons in a part of AMSOM: $N_u$ is highlighted as the neuron with the highest error and $N_v$ is the neuron among its neighbors with the largest error. Neurons $N_1$ and $N_2$ are added instead of $N_u$, matrices $E$ and $A$ are updated accordingly and weight/position vectors are determined by Equations \ref{eqn:neuronadd1}-\ref{eqn:neuronadd4}.}
  \label{fig:msomadd}
\end{figure}

Regarding the new neuron that will be added, we follow the the biological process of \lq cell division\rq\ \cite{odri1993evolutional}. By this way the neuron with the highest quantization error is \lq splitted\rq\ to two new neurons (instead of just adding one new neuron somewhere at random with no connections at all). Both new neurons preserve the same connectivity (and also they are connected to each other) with the original neuron, thus we achieve a preservation of behavioral link between the parent and the offspring. Regarding the exact position of the two neurons the following process is followed: Neuron with the largest error among $N_u$'s neighbors is spotted (let it be $N_v$). One neuron will preserve $N_u$'s position and the other one will be placed in the middle between $N_u$ and $N_v$. In detail, weights and positions of the two new neurons ($u_1$ and $u_2$) are calculated using the following Equations:
\begin{equation}
\mathbf{w}_{u1} = (1+\beta) \times \mathbf{w}_u
\label{eqn:neuronadd1}
\end{equation}
\begin{equation}
\mathbf{w}_{u2} = -\beta \times \mathbf{w}_u
\label{eqn:neuronadd2}
\end{equation}
\begin{equation}
\mathbf{r}_{u1} = \mathbf{r}_{u}
\label{eqn:neuronadd3}
\end{equation}
\begin{equation}
\mathbf{r}_{u2} = \frac{\mathbf{r}_u+\mathbf{r}_v}{2}
\label{eqn:neuronadd4}
\end{equation}

where $\mathbf{w}_{u}$ refers to the weight vector of neuron $u$ (neuron that is splitted) and $\beta$ is a mutation parameter which can take either a fixed or random value according to a certain distribution rule (following \cite{odri1993evolutional}). In any case, value of $\beta$ has to be chosen small in order to avoid a large change both in network topology but also in the weight vectors. In this paper, $\beta$ takes a random value according to a Gaussian distribution with a mean of zero and variance of one. New neurons retain the same connectivity to other neurons as the parent neuron but age weights are zeroed. The process of adding a new neuron (along with any changes in matrices $E$ and $A$) is described in Figure \ref{fig:msomadd}.

It has to be pointed out that there is the possibility that a neuron would be removed from a region of the map and to be added in another region (removing and adding neurons are consecutive processes). This comes to agreement with several theories in neural organization, suggesting that cortical regions can adapt to their input sources and are somewhat interchangeable or \lq reusable\rq\ by other modalities, especially in vision- or hearing-impaired subjects \cite{wedeen2012geometric}.

\subsubsection{Architecture Adaptation and Termination Criterion}
\begin{figure}[htb]
\centering
  \includegraphics[width=1.0\linewidth]{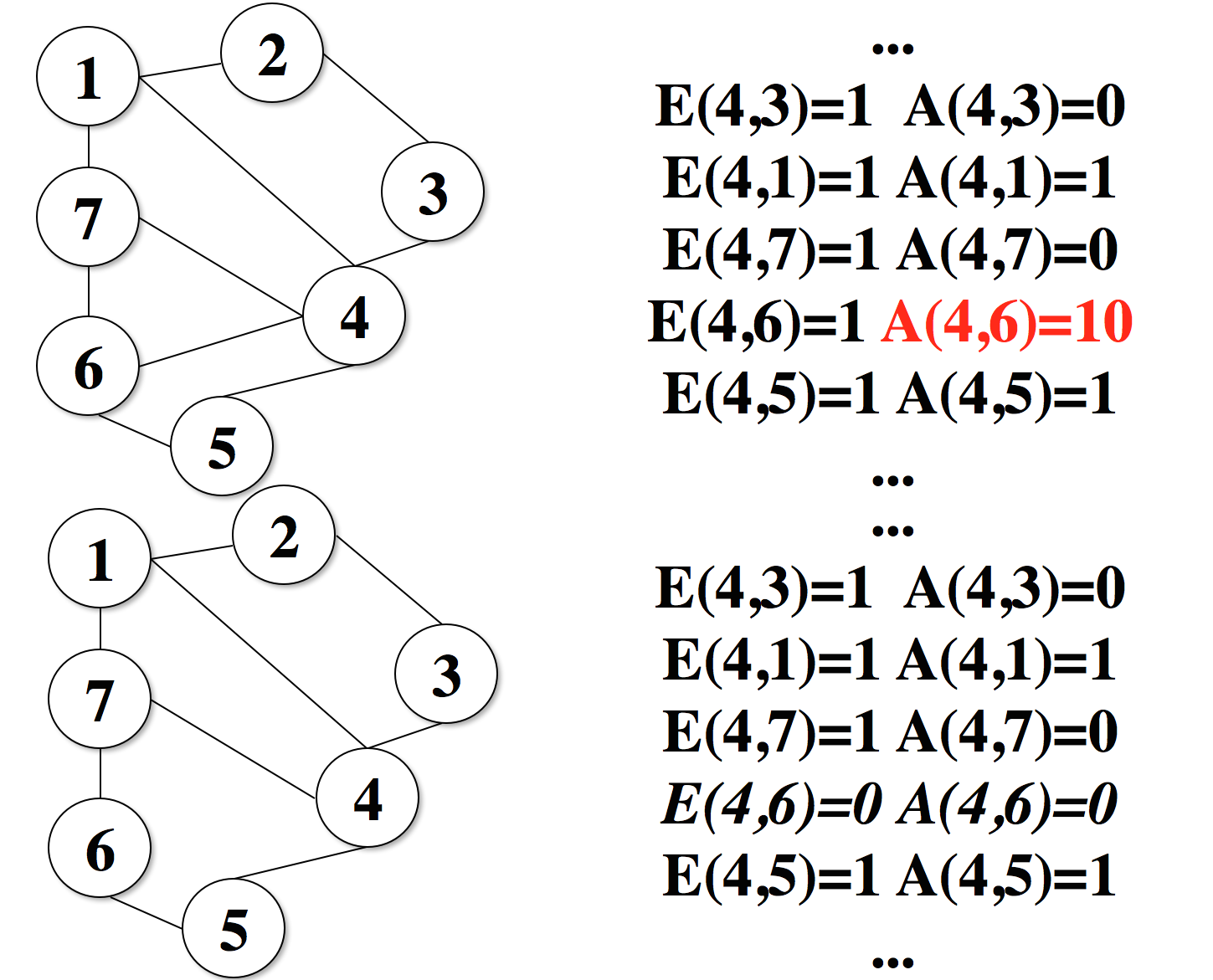}
  \caption{Maintaining the structure of AMSOM: With $Q=4$ (i.e., a rectangular grid) neuron 4 is connected to five neurons, so it's connection with neuron 6 (oldest connection) is removed.}
  \label{fig:msomadapt}
\end{figure}

As it is described before, initial structure of AMSOM is adapted through learning and training in order to find what is optimal for the number of neurons, their weights and their connections. The adaptation process starts by training the initial structure of AMSOM. When the criteria of adding or removing neurons are satisfied, then the network is adapted. In order to maintain (as possible) the initial structure (i.e., rectangular or hexagonal or any other lattice selected), after this adaptation process we re-evaluate all connections of all neurons and make sure that each neuron has at most $Q$ neighbors (where $Q$ is decided in the beginning, i.e., in the case of rectangular lattice, $Q=4$): This can be ensured by checking edge matrix $E$ after each epoch and if a neuron is found to have more than $Q$ connections then only the $Q$-\lq recent\rq\ are kept (utilizing age of edges in matrix $A$). This process is presented in Figure \ref{fig:msomadapt}.

By this training scheme, AMSOM adapts simultaneously the structure of the map (number of neurons and connections) and the weight vectors. Removing and adding neurons occur when different criteria are met, so they can be applied in any sequence, depending on when the criteria are satisfied. By applying these operations repeatedly, AMSOM is expected to find a near-optimal structure and representation of a given dataset.

Finally, like every SOM algorithm, AMSOM has an upper limit of epochs that training takes place. This number is set to 1000 but there is also a premature termination criterion depending on the mean quantization error change between two consecutive epochs. Thus, if $mqe(t)-mqe(t-1) < \epsilon_1$ where $\epsilon_1$ is a small value (like $1E-06$) then the map has reached the desired size (according to the $GT$ provided) and training is terminated.

\subsection{Phase III: AMSOM Finalization}
Final phase of AMSOM happens when learning is complete and structure of the network is not any more changing. No neurons are added or removed at this phase and no connections between neurons are added or removed but weight and position vector adaptation is continued with a lower rate. Purpose of this process is to smooth out any quantization error and fine tune weights and positions of the neurons, especially for neurons added at the latter epochs. For this purpose, neighborhood function (both for Equations \ref{eqn:msomtrain} and \ref{eqn:posupd} is constrained only to the immediate neighborhood and learning rate $\alpha(t)$ in Equation (\ref{eqn:posupd}) is set to 0.001 (even smaller than in phase II). Phase III is concluded when there is no significant change in change in mean quantization error (i.e., when $mqe(t) - mqe(t-1) < \epsilon_2$), where $\epsilon_2$ is set to a smaller value than $\epsilon_1$ (like $1E-10$).

\section{\uppercase{Experiments}}
\label{sect:exp}

AMSOM performance has been tested with several literature datasets in order to evaluate both map quality (in terms of topology preservation) and the number of epochs needed to converge. Quantization Error (QE) and Topographic Error (TE) were used as intrinsic measures of evaluation (for more details readers are encouraged to read \cite{bauer1999neural}). All datasets were provided by the UCI repository \footnote{http://archive.ics.uci.edu/ml/}, except the CLUSTER dataset which is a simple and random but large two-dimensional dataset with four groups. All datasets used with their characteristics are presented in Table \ref{tab:datasets}.

\begin{table}[htb]
  \centering
  \caption{Datasets characteristics.}
    \begin{tabular}{cccc}
    \hline
    \textbf{Dataset name} & \textbf{Instances} & \textbf{Features} & \textbf{Classes} \\
    \hline
    CLUSTER & 1000  & 2     & 4 \\
    IRIS  & 150   & 4     & 3 \\
    WINE  & 178   & 13    & 7 \\
    IONOSPHERE & 351   & 35    & 2 \\
    CANCER & 699   & 9     & 2 \\
    GLASS  & 214   & 10     & 7 \\
    \hline
    \end{tabular}%
  \label{tab:datasets}%
\end{table}

\begin{table*}[htbp]
  \centering
  \caption{Quality of AMSOM compared to classic SOM and number of neurons.}
    \begin{tabular}{rcccccc}
\hline
          & \multicolumn{2}{c}{QE} & \multicolumn{2}{c}{TE} & \multicolumn{2}{c}{Number of neurons} \\
    \hline
    \multicolumn{1}{c}{\textbf{Dataset name}} & \textbf{AMSOM} & \textbf{SOM} & \textbf{AMSOM} & \textbf{SOM} & \textbf{AMSOM} & \textbf{SOM} \\
    \multicolumn{1}{c}{CLUSTER} & 0.108 & 0.1090 & 0.028 & 0.063 & 121   & 154 \\
    \multicolumn{1}{c}{IRIS} & 0.1047 & 0.3930 & 0.009 & 0.013 & 37    & 66 \\
    \multicolumn{1}{c}{WINE} & 1.7394 & 1.8830 & 0.008 & 0.017 & 42    & 66 \\
    \multicolumn{1}{c}{IONOSPHERE} & 2.5697 & 2.9418 & 0.0026 & 0.0057 & 78    & 91 \\
    \multicolumn{1}{c}{CANCER} & 0.7941 & 0.9456 & 0.0145 & 0.0286 & 103   & 132 \\
    \multicolumn{1}{c}{GLASS} & 0.9797 & 1.1178 & 0.0041 & 0.0093 & 43    & 72 \\
    \hline
    \end{tabular}%
  \label{tab:results}%
\end{table*}

\begin{figure*}[htb]
  \centering
 \subfigure[Neuron grid]{
        \includegraphics[width=\columnwidth,keepaspectratio]{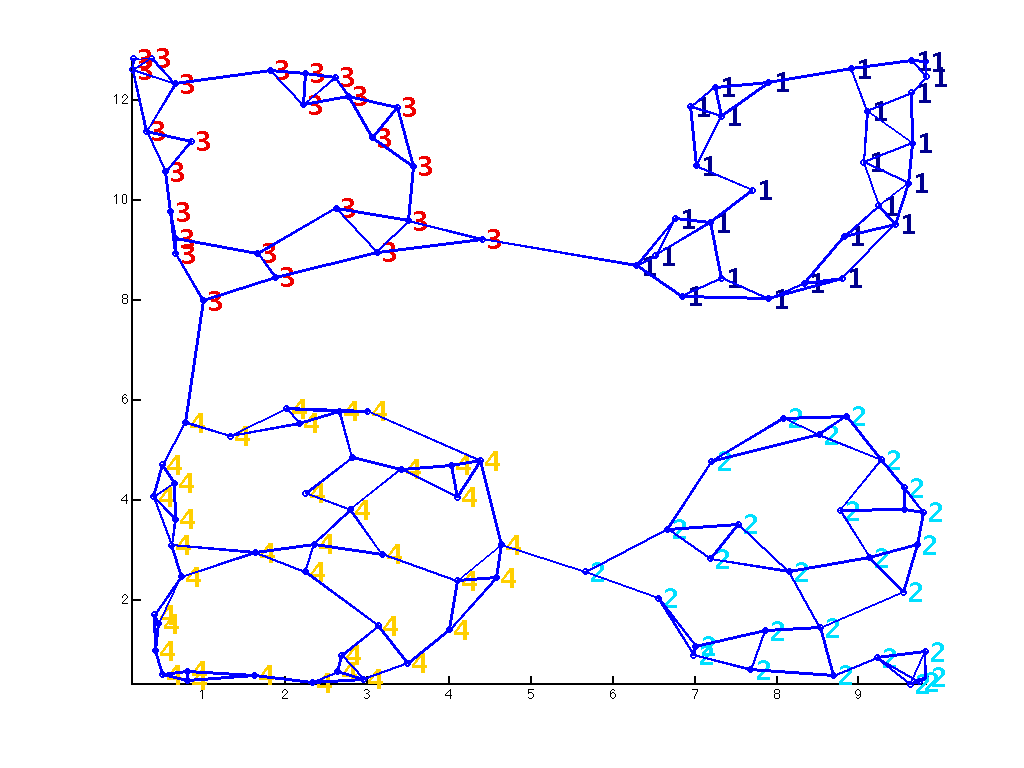}}
 \subfigure[Neuron classes]{
        \includegraphics[width=\columnwidth,keepaspectratio]{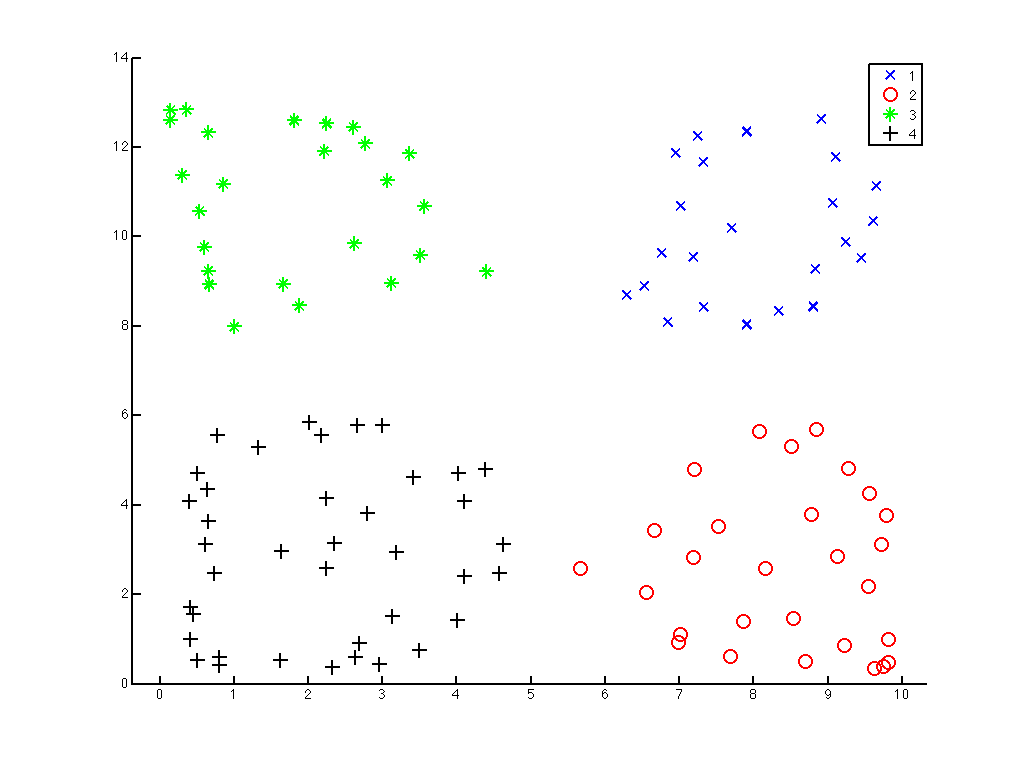}}
\caption{Visualization results for CLUSTER dataset (4 classes).}
\label{fig:cluster}
\end{figure*}

\begin{figure*}[htb]
  \centering
  \subfigure[Neuron grid]{
        \includegraphics[width=\columnwidth,keepaspectratio]{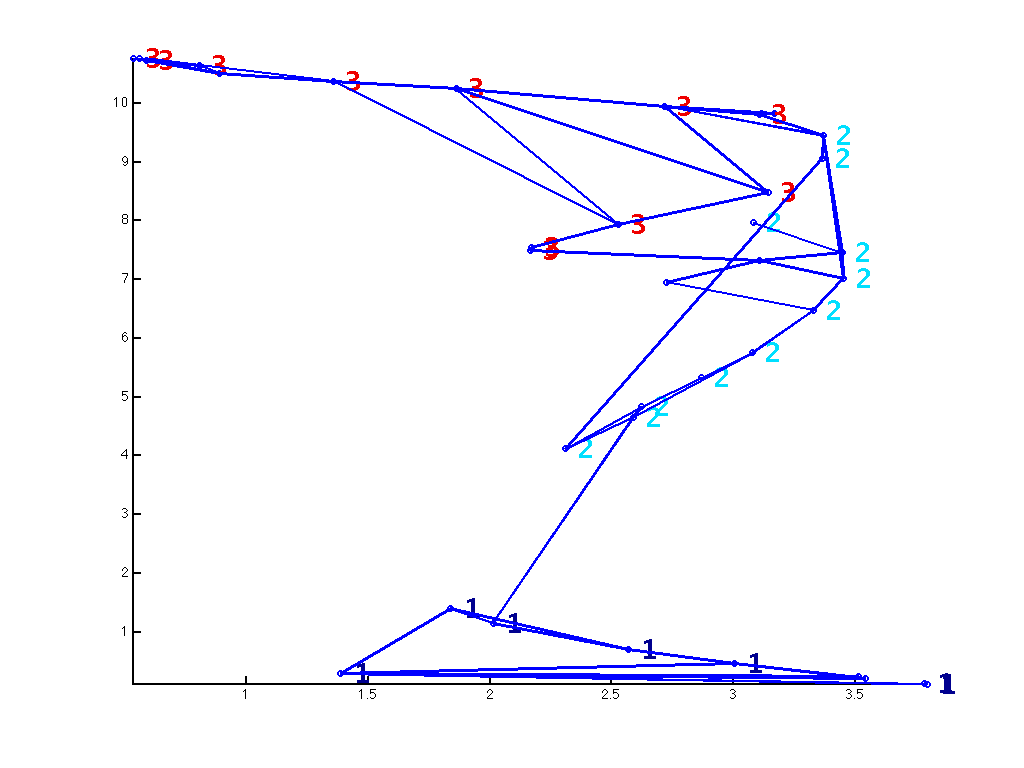}}
  \subfigure[Neuron classes]{
        \includegraphics[width=\columnwidth,keepaspectratio]{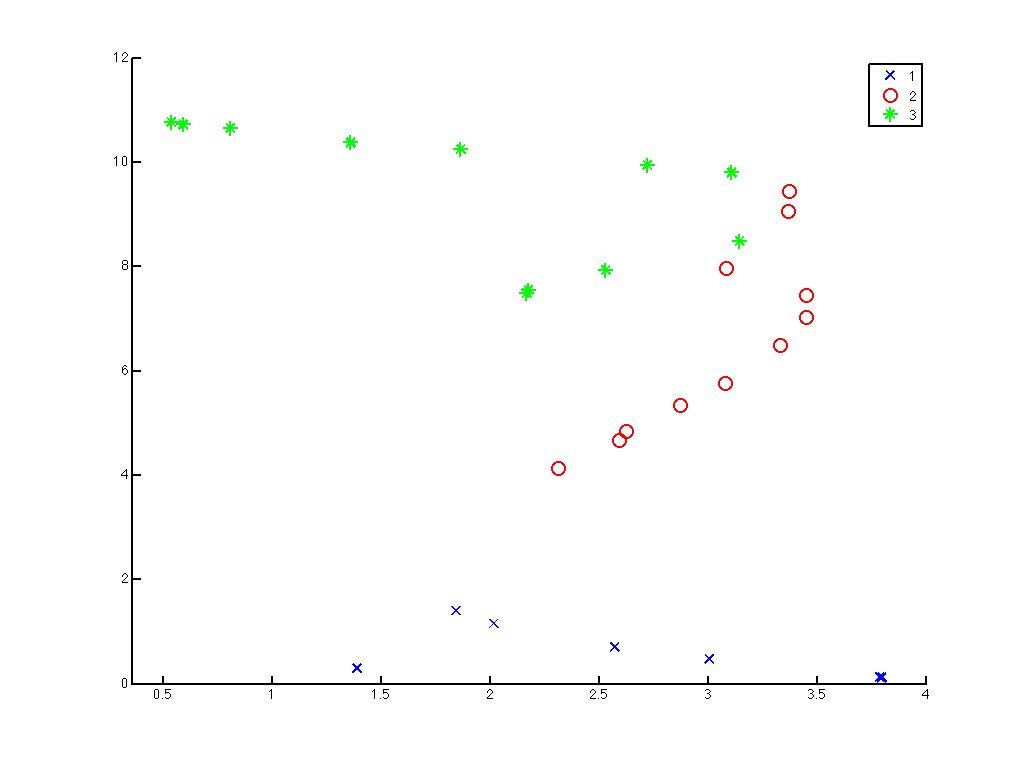}}
\caption{Visualization results for IRIS dataset (3 classes).}
\label{fig:iris}
\end{figure*}

\begin{figure*}[htb]
  \centering
  \subfigure[Neuron grid]{
        \includegraphics[width=\columnwidth,keepaspectratio]{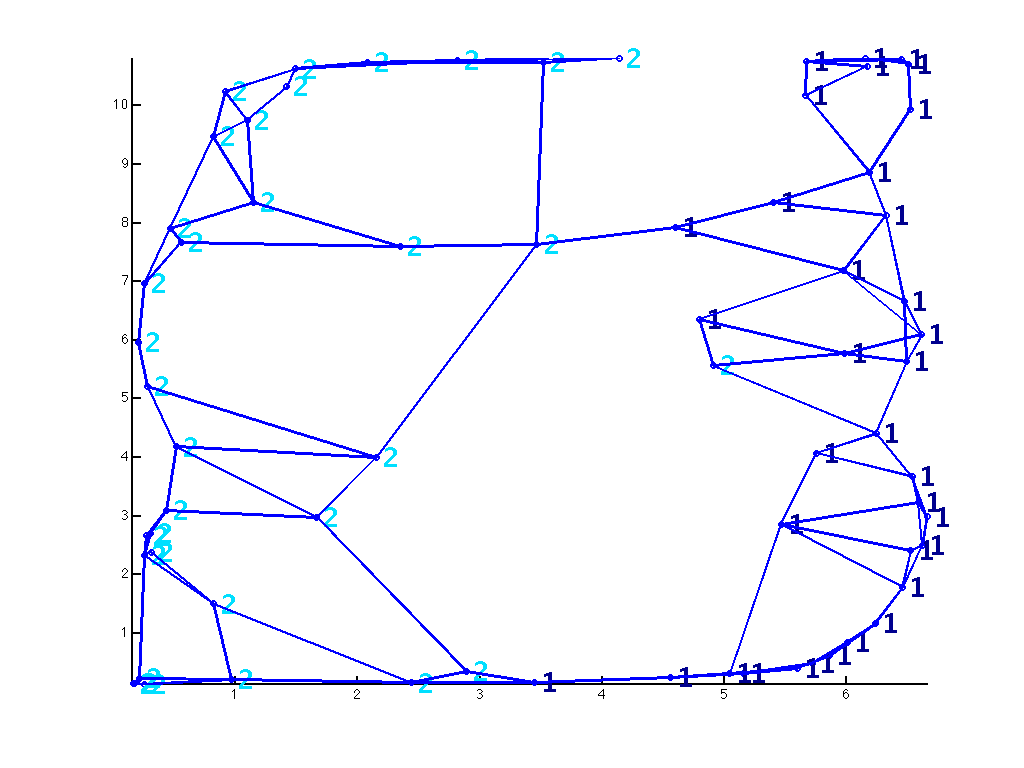}}
  \subfigure[Neuron classes]{
        \includegraphics[width=\columnwidth,keepaspectratio]{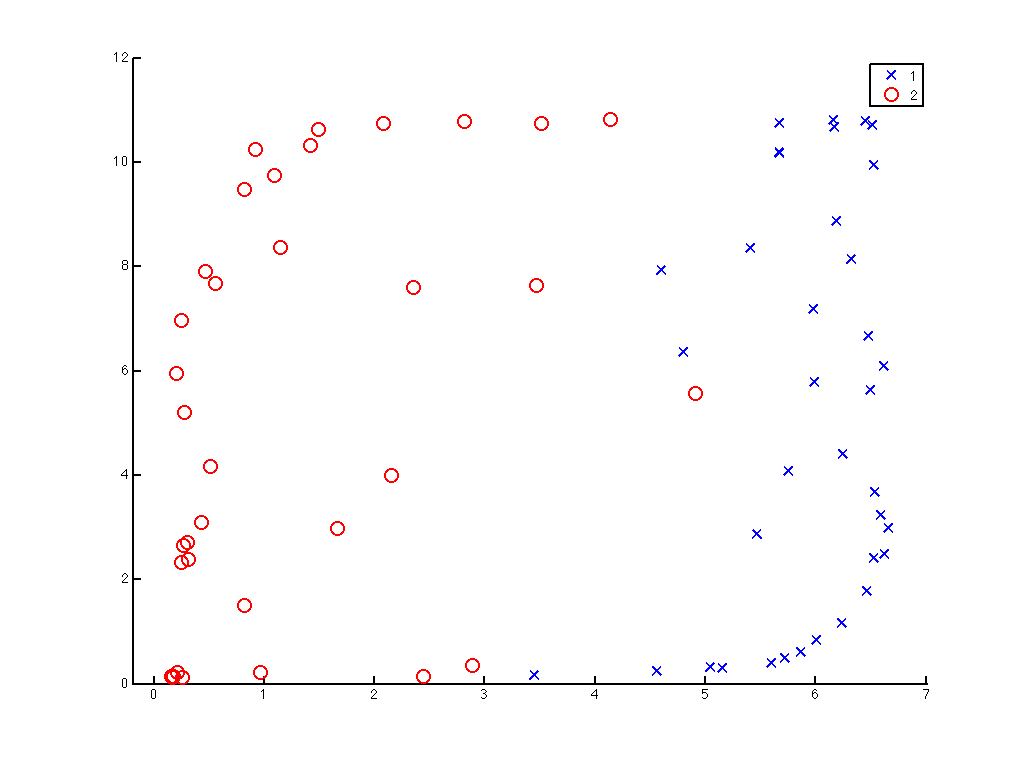}}
\caption{Visualization results for IONOSPHERE dataset (2 classes).}
\label{fig:ionosphere}
\end{figure*}

Each dataset in shuffled and split to training, testing and validation set (60\%, 20\% and 20\% respectively). Each experiment was performed 20 times and the results presented here are average over these runs (deviations were small and are not presented here). Results for AMSOM QE and TE (compared to classic SOM) along with the number of neurons used by each model are presented in Table \ref{tab:results}. From this Table it is obvious that AMSOM's performance is much better than classic SOM. AMSOM starts from the same number of neurons as classic SOM but by removing and adding neurons when necessary reaches a number which is suitable to represent the dataset. Both QE and TE are improved using AMSOM algorithm and this improvement is more significant in TE because of the neuron position changing which allows the map to better adjust to the dataset.

Visualization results for three of the datasets are presented in Figures \ref{fig:cluster} through \ref{fig:ionosphere}. In these figures final positions of the neurons and their final positions are represented. For each neuron a majority vote criterion was used in order to determine the class that this neuron represents. For the simple CLUSTER dataset it is obvious that the four classes are identified and the grid structure can effectively represent their relations. For the IRIS dataset one class is completely identified whereas the other two (which are more similar) are also highlighted. Also notice that neurons that belong to the same class are mostly connected with each other on the grid and only some spontaneous connections between classes exist. Finally, for the more demanding IONOSPHERE dataset (see the relatively higher QE), AMSOM manages to differentiate in a great degree the two classes. Neuron grids in all figures also reveal that the percentage of dead units (neurons that do not represent any pattern) is significantly small, which is an improvement to the classic SOM algorithm (fewer inactive neurons).

Regarding the spread factor ($SF$) which controls the growing threshold ($GT$), a value $0.5$ was chosen for this series of experiments because for all datasets it yielded satisfactory results. In the general case that there is no prior knowledge on the data examined, a low value of $SF$ (0-0.3) will allow highlighting of the most significant clusters. 
Regarding $\gamma$ parameter of Equation (\ref{eqn:posupd}) it was found that it can effectively control the spreading or shrinking of neighborhood during position updating and by this way creating more isolated or more connected clusters. Several experiments were conducted (not presented here due to space limitations) and showed that small values of gamma (1 till 10) produce the best results for all datasets. The higher the $\gamma$, the better topographic preservation (reduced TE) but the quantization error (QE) rises. Also, high values of $\gamma$ tend to increase the number of neurons that remain unused (dead units) whereas values close to 100 tend to approach the classic SOM algorithm (position updating is minimal).
Two more parameters that need to be adjusted are $age_{max}$ and $t_{add}$. For both parameters, 30 epochs were found to be optimal, which is sound given the fact that 30 epochs are enough time to see if current structure performs well (reduced QE) or if adjustments needed (adding/removing neurons).

Complexity of the developed algorithm is slightly increased due to the need for updating matrices $A$ and $E$ and also due to the more flexible structure. This overhead is partly counterbalanced by the faster training process (in all experiments there was a decrease in epochs number around 20\%) since updating neuron positions clearly improves training time (requires less epochs) but for memory intensive tasks (like big datasets) this might be become a drawback for the algorithm.

\section{\uppercase{Conclusion}}
\label{sect:concl}
In this paper we presented AMSOM, an extension to original SOM algorithm which allows neurons to change positions according to a similar competitive technique used in classic SOM training. Moreover, neurons can be added or removed during this \quotes{double} training process allowing for a more flexible structure grid which is able to represent the dataset more efficiently.  Experimental results on different datasets improve performance of AMSOM compared to classic SOM algorithm. AMSOM produces better reference vectors by reducing the quantization error, topology is preserved through the neuron moving by significantly reducing the Topographic Error and the visualization result matches as much as possible the original dataset partitions. Also, AMSOM produces fewer nodes with no significant effect while at the same time it reduces required number of epochs.

Obtained results give new insights on how to utilize the concept of competitive learning and self-organization in neural networks and will be examined in more detail so as to further improve performance and investigate behavior in bigger and real-life datasets (images, text, etc.). For this purpose, drawbacks of the algorithm (the need to tune four parameters and the memory cost) will be addressed in order to further facilitate visualization and clustering (especially in large datasets).

\vfill
\bibliography{som}

\begin{thebibliography}{}

\bibitem[Alahakoon et~al., 2000]{alahakoon2000dynamic}
Alahakoon, D., Halgamuge, S.~K., and Srinivasan, B. (2000).
\newblock Dynamic self-organizing maps with controlled growth for knowledge
  discovery.
\newblock {\em Neural Networks, IEEE Transactions on}, 11(3):601--614.

\bibitem[Ayadi et~al., 2012]{ayadi2012migsom}
Ayadi, T., Hamdani, T.~M., and Alimi, A.~M. (2012).
\newblock {MIGSOM: multilevel interior growing self-organizing maps for high
  dimensional data clustering}.
\newblock {\em Neural processing letters}, 36(3):235--256.

\bibitem[Bauer et~al., 1999]{bauer1999neural}
Bauer, H.-U., Herrmann, M., and Villmann, T. (1999).
\newblock Neural maps and topographic vector quantization.
\newblock {\em Neural networks}, 12(4):659--676.

\bibitem[Blackmore and Miikkulainen, 1993]{blackmore1993incremental}
Blackmore, J. and Miikkulainen, R. (1993).
\newblock {Incremental grid growing: encoding high-dimensional structure into a
  two-dimensional feature map}.
\newblock In {\em Neural Networks, 1993., IEEE International Conference on},
  pages 450--455.

\bibitem[Bortman and Aladjem, 2009]{bortman2009growing}
Bortman, M. and Aladjem, M. (2009).
\newblock A growing and pruning method for radial basis function networks.
\newblock {\em Neural Networks, IEEE Transactions on}, 20(6):1039--1045.

\bibitem[Ceccarelli et~al., 1993]{ceccarelli1993competitive}
Ceccarelli, M., Petrosino, A., and Vaccaro, R. (1993).
\newblock Competitive neural networks on message-passing parallel computers.
\newblock {\em Concurrency: Practice and Experience}, 5(6):449--470.

\bibitem[Changeux and Danchin, 1976]{Changeux}
Changeux, J.-P. and Danchin, A. (1976).
\newblock Selective stabilisation of developing synapses as a mechanism for the
  specification of neuronal networks.
\newblock {\em Nature}, 264(5588):705--712.

\bibitem[Deboeck and Kohonen, 2013]{deboeck2013visual}
Deboeck, G. and Kohonen, T. (2013).
\newblock {\em Visual explorations in finance: with self-organizing maps}.
\newblock Springer Science \& Business Media.

\bibitem[Doherty et~al., 2005]{doherty2005treegng}
Doherty, K., Adams, R., and Davey, N. (2005).
\newblock {TreeGNG-hierarchical topological clustering}.
\newblock In {\em ESANN}, pages 19--24.

\bibitem[Dowling, 2007]{dowling2007great}
Dowling, J.~E. (2007).
\newblock {\em The Great Brain Debate: Nature Or Nurture?}
\newblock Princeton University Press.

\bibitem[Est{\'e}vez et~al., 2012]{estevez2012advances}
Est{\'e}vez, P.~A., Pr{\'\i}ncipe, J.~C., and Zegers, P. (2012).
\newblock {\em Advances in Self-Organizing Maps: 9th International Workshop,
  WSOM 2012 Santiago, Chile, December 12-14, 2012 Proceedings}.
\newblock Springer Science \& Business Media.

\bibitem[Fort, 2006]{fort2006som}
Fort, J.-C. (2006).
\newblock Som’s mathematics.
\newblock {\em Neural Networks}, 19(6):812--816.

\bibitem[Fritzke, 1994]{fritzke1994growing}
Fritzke, B. (1994).
\newblock Growing cell structures—a self-organizing network for unsupervised
  and supervised learning.
\newblock {\em Neural networks}, 7(9):1441--1460.

\bibitem[Fritzke, 1995]{fritzke1995gg}
Fritzke, B. (1995).
\newblock Growing grid—a self-organizing network with constant neighborhood
  range and adaptation strength.
\newblock {\em Neural Processing Letters}, 2(5):9--13.

\bibitem[Fritzke et~al., 1995]{fritzke1995growing}
Fritzke, B. et~al. (1995).
\newblock A growing neural gas network learns topologies.
\newblock {\em Advances in neural information processing systems}, 7:625--632.

\bibitem[Han and Qiao, 2013]{han2013structure}
Han, H.-G. and Qiao, J.-F. (2013).
\newblock A structure optimisation algorithm for feedforward neural network
  construction.
\newblock {\em Neurocomputing}, 99:347--357.

\bibitem[Hodge and Austin, 2001]{hodge2001hierarchical}
Hodge, V.~J. and Austin, J. (2001).
\newblock {Hierarchical growing cell structures: TreeGCS}.
\newblock {\em Knowledge and Data Engineering, IEEE Transactions on},
  13(2):207--218.

\bibitem[Islam et~al., 2009]{islam2009new}
Islam, M., Sattar, A., Amin, F., Yao, X., and Murase, K. (2009).
\newblock A new adaptive merging and growing algorithm for designing artificial
  neural networks.
\newblock {\em Systems, Man, and Cybernetics, Part B: Cybernetics, IEEE
  Transactions on}, 39(3):705--722.

\bibitem[Kohonen, 1981]{kohonen1981automatic}
Kohonen, T. (1981).
\newblock Automatic formation of topological maps of patterns in a
  self-organizing system.

\bibitem[Kohonen, 1988]{kohonen1988neural}
Kohonen, T. (1988).
\newblock The'neural'phonetic typewriter.
\newblock {\em Computer}, 21(3):11--22.

\bibitem[Kohonen, 1993]{kohonen1993things}
Kohonen, T. (1993).
\newblock {Things you haven't heard about the Self-Organizing Map}.
\newblock In {\em Neural Networks, 1993., IEEE International Conference on},
  pages 1147--1156. IEEE.

\bibitem[Kohonen, 2001]{kohonen2001self}
Kohonen, T. (2001).
\newblock {\em {Self-organizing Maps, vol. 30 of Springer Series in Information
  Sciences}}.
\newblock Springer Berlin.

\bibitem[Kohonen, 2012]{kohonen2012self}
Kohonen, T. (2012).
\newblock {\em Self-organization and associative memory}, volume~8.
\newblock Springer.

\bibitem[Lagus et~al., 1999]{lagus1999websom}
Lagus, K., Honkela, T., Kaski, S., and Kohonen, T. (1999).
\newblock {WEBSOM for textual data mining}.
\newblock {\em Artificial Intelligence Review}, 13(5-6):345--364.

\bibitem[Lu, 1990]{lu1990pattern}
Lu, S.-y. (1990).
\newblock Pattern classification using self-organizing feature maps.
\newblock In {\em 1990 IJCNN International Joint Conference on}, pages
  471--480.

\bibitem[Marsland et~al., 2002]{marsland2002self}
Marsland, S., Shapiro, J., and Nehmzow, U. (2002).
\newblock A self-organising network that grows when required.
\newblock {\em Neural Networks}, 15(8):1041--1058.

\bibitem[Mulier and Cherkassky, 1994]{mulier1994learning}
Mulier, F. and Cherkassky, V. (1994).
\newblock Learning rate schedules for self-organizing maps.
\newblock In {\em Pattern Recognition, 1994. Vol. 2-Conference B: Computer
  Vision \&amp; Image Processing., Proceedings of the 12th IAPR International.
  Conference on}, volume~2, pages 224--228. IEEE.

\bibitem[Odri et~al., 1993]{odri1993evolutional}
Odri, S.~V., Petrovacki, D.~P., and Krstonosic, G.~A. (1993).
\newblock Evolutional development of a multilevel neural network.
\newblock {\em Neural Networks}, 6(4):583--595.

\bibitem[Park et~al., 2006]{Park2006247}
Park, Y.-S., Tison, J., Lek, S., Giraudel, J.-L., Coste, M., and Delmas, F.
  (2006).
\newblock {Application of a self-organizing map to select representative
  species in multivariate analysis: A case study determining diatom
  distribution patterns across France}.
\newblock {\em Ecological Informatics}, 1(3):247 -- 257.
\newblock 4th International Conference on Ecological Informatics.

\bibitem[Rauber et~al., 2002]{rauber2002growing}
Rauber, A., Merkl, D., and Dittenbach, M. (2002).
\newblock The growing hierarchical self-organizing map: exploratory analysis of
  high-dimensional data.
\newblock {\em Neural Networks, IEEE Transactions on}, 13(6):1331--1341.

\bibitem[Spanakis et~al., 2012]{spanakis2012doso}
Spanakis, G., Siolas, G., and Stafylopatis, A. (2012).
\newblock {DoSO: a document self-organizer}.
\newblock {\em Journal of Intelligent Information Systems}, 39(3):577--610.

\bibitem[Vesanto et~al., 2000]{vesanto2000som}
Vesanto, J., Himberg, J., Alhoniemi, E., and Parhankangas, J. (2000).
\newblock {\em {SOM toolbox for Matlab 5}}.
\newblock Citeseer.

\bibitem[Wedeen et~al., 2012]{wedeen2012geometric}
Wedeen, V.~J., Rosene, D.~L., Wang, R., Dai, G., Mortazavi, F., Hagmann, P.,
  Kaas, J.~H., and Tseng, W.-Y.~I. (2012).
\newblock The geometric structure of the brain fiber pathways.
\newblock {\em Science}, 335(6076):1628--1634.

\bibitem[Yang and Chen, 2012]{yang2012evolutionary}
Yang, S.-H. and Chen, Y.-P. (2012).
\newblock An evolutionary constructive and pruning algorithm for artificial
  neural networks and its prediction applications.
\newblock {\em Neurocomputing}, 86:140--149.

\end{thebibliography}

\end{document}